\documentclass[journal,onecolumn]{IEEEtran}
\usepackage{array}
\usepackage{amsmath}
\usepackage{mathrsfs}
\usepackage[utf8]{inputenc} % allow utf-8 input
\usepackage[T1]{fontenc}    % use 8-bit T1 fonts
\usepackage{hyperref}       % hyperlinks
\usepackage{url}            % simple URL typesetting
\usepackage{booktabs}       % professional-quality tables
\usepackage{amsfonts}       % blackboard math symbols
\usepackage{nicefrac}       % compact symbols for 1/2, etc.
\usepackage{microtype}      % microtypography
\usepackage{lipsum}
\usepackage{graphicx}
\graphicspath{ {./images/} }
\usepackage{blindtext}
\usepackage[inline]{enumitem}
\usepackage[dvipsnames]{xcolor}
\usepackage{float}
\usepackage{multirow}
\begin{document}
%\setpagewiselinenumbers

\title{Spatiotemporal Predictions of Toxic Urban Plumes Using Deep Learning}

\author{Yinan Wang,~\IEEEmembership{}
        M. Giselle Fernández-Godino,~\IEEEmembership{}
        Nipun Gunawardena,~\IEEEmembership{}
        Donald D. Lucas,~\IEEEmembership{}
        and~Xiaowei Yue~\IEEEmembership{}
        \IEEEcompsocitemizethanks{

\IEEEcompsocthanksitem Yinan Wang is with the
Department of Industrial and Systems Engineering, Rensselaer Polytechnic Institute, Troy, NY, 12180.
E-mail: wangy88@rpi.edu
\IEEEcompsocthanksitem M. Giselle Fernández-Godino, Nipun Gunawardena, and Donald D. Lucas are with Lawrence Livermore National Laboratory, Livermore, CA, 94550.
E-mail: \{fernandez48, gunawardena1, lucas26\}@llnl.gov
\IEEEcompsocthanksitem Xiaowei Yue is with the Department of Industrial Engineering and the Institute for Quality and Reliability, Tsinghua University, Beijing, China, 100084. E-mail: yuex@mail.tsinghua.edu.cn% 
}
 % stops an unwanted space
% \thanks{This work was performed under the auspices of the United States Department of Energy (DOE) by Lawrence Livermore National Laboratory under contract DE-AC52-07NA27344.}
% \thanks{This paper has supplementary downloadable material available at http://ieeexplore.ieee.org, provided by the authors. This includes eight multimedia GIF format video clips, which demonstrate the performance comparison on the testing dataset. This material is 131 KB in size.}
% \thanks{Manuscript received xxxx; revised xxxx.}
% \thanks{(Corresponding author: Xiaowei Yue, e-mail: xwy@vt.edu, phone: 540-231-9081, fax: 540-231-3322)}
}

\markboth{Preprint}%
{Shell \MakeLowercase{\textit{et al.}}: Bare Demo of IEEEtran.cls for Computer Society Journals}




\IEEEtitleabstractindextext{%

\begin{abstract}
Industrial accidents, chemical spills, and structural fires can release large amounts of harmful materials that disperse into urban atmospheres and impact populated areas. Computer models are typically used to predict the transport of toxic plumes by solving fluid dynamical equations. However, these models can be computationally expensive due to the need for many grid cells to simulate turbulent flow and resolve individual buildings and streets. In emergency response situations, alternative methods are needed that can run quickly and adequately capture important spatiotemporal features. Here, we present a novel deep learning model called \emph{ST-GasNet} that was inspired by the mathematical equations that govern the behavior of plumes as they disperse through the atmosphere. ST-GasNet learns the spatiotemporal dependencies from a limited set of temporal sequences of ground-level toxic urban plumes generated by a high-resolution large eddy simulation model. On independent sequences, ST-GasNet accurately predicts the late-time spatiotemporal evolution, given the early-time behavior as an input, even for cases when a building splits a large plume into smaller plumes. By incorporating large-scale wind boundary condition information, ST-GasNet achieves a prediction accuracy of at least 90\% on test data for the entire prediction period.
\end{abstract}

% Note that keywords are not normally used for peer review papers.
}

\maketitle

\IEEEdisplaynontitleabstractindextext

% \fi
\IEEEpeerreviewmaketitle

\section{Introduction}
\label{sec:introduction}
%\linenumbers
\IEEEPARstart{M}any harmful airborne materials are dispersed into the atmosphere from human activities. Many of these materials are introduced in low quantities from the daily operation of coal-fired power plants, industrial processes, refineries, and vehicles. Some, however, result from accidental events, and their concentration is often high, posing a threat to life in urban areas. Industrial or residential fires, explosions, and gas leaks are examples of such accidents that can produce toxic plumes. The dispersion of these materials might pose severe environmental harm and can even threaten human lives \cite{Griffin2004, MORENO2007913, atmos12080953}. We refer to {\it toxic plumes} as air contamination caused by unexpected or sudden releases of airborne materials and use the terms \textit{evolution} or \textit{dispersion} to describe their time-dependent behavior due to atmospherically-driven advection and turbulent mixing. Because toxic plumes in urban environments can reach populated areas quickly, fast predictions can help decision-makers and emergency responders, and help reduce associated socioeconomic losses. Unlike air quality pollutants that are more ubiquitous (e.g., ozone and carbon monoxide), toxic plumes occur less often, which can make it challenging to predict their evolution. Furthermore, it is usually more difficult to predict or simulate dispersion in urban regions than in rural areas due to the presence of buildings, street canyons, and other terrain features that enhance atmospheric turbulence.

% For predictions of 9 minutes ahead, with the wind speeds 
% condsidered in this paper, the plume travels more than hundreds of meters.

Physics-based simulations aim to solve the problem of plume evolution by discretizing the domain and solving the differential equations that govern it. One standard approach is the Gaussian plume model, which is used to study the transport of airborne contaminants \cite{Stockie2011TheMO}. The mass continuity equation in Equation~\eqref{eq:AdvDiff} describes the spatiotemporal evolution of contaminant concentration $C(\mathbf{x},t)$ in three dimensions subject to advection and diffusion:
\begin{equation}\label{eq:AdvDiff}
\frac{\partial C(\mathbf{x},t)}{\partial t} + \nabla \cdot \left( \vec{u} C(\mathbf{x},t) \right) = \left( \nabla \cdot \mathbf{K} \nabla \right) C(\mathbf{x},t) + S(\mathbf{x},t),
\end{equation}
here, $\mathbf{x}$ represents location in $x$, $y$, and $z$; $t$ represents time; $\vec{u}$ is the three-dimensional wind vector; $\mathbf{K}$ is the eddy diffusion tensor; $\nabla$ is the differential operator in $\mathbf{x}$; and $S(\mathbf{x},t)$ is the source or sink term. Analytical solutions to this equation can be obtained under restrictive assumptions, leading to the Gaussian plume class of models \cite{Stockie2011TheMO}. Although Gaussian plume models are computationally efficient, they are also unsuitable in urban environments where the wind fields are turbulent and unsteady~\cite{bady2006comparative}. To better resolve unsteady fluid flow, computational fluid dynamics (CFD) codes are used to solve the Navier-Stokes equations using a variety of approaches, including direct numerical simulations~\cite{moin1998direct}, large eddy simulations~\cite{zhiyin2015large}, or Reynolds Averaged Navier Stokes simulations~\cite{alfonsi2009reynolds}. The resulting CFD flow fields can then be used for contaminant transport in Equation~\eqref{eq:AdvDiff}. However, higher fidelity winds come at an increased computational cost.

% I recommend removing the Lagrangian equation in the introduction for clarification. We should discuss the transport algorithm in later in section 3. 
%The equations for a Lagrangian particle displacement $x_i(t)$ due to advection, diffusion, and settling in the three coordinate directions $i=x, y, z$ are:
%
%\begin{equation}\label{eq:AdvDiff2}
%\partial x_i = \tilde{u}_i dt +  \frac{\frac{ \partial k_i}{S_c}}{ d %x_i } dt + \frac{2 k_i}{S_c} d W_{x_i},
%\end{equation}
%%
%where $\tilde{u}_i$ represents the wind components and $k_i$ the diffusion coefficient in the $x$, $y$, and $z$ directions; $S_c$ is the Schmidt number; $d W{x_i}$ are three independent normal random variates with zero mean and variance $dt$, with $dt$ being the advection timestep of a Lagrangian particle. At any time $t$, the concentration can be calculated from the Lagrangian particle locations and concentration.
\begin{figure*}[!ht]
    \centering
    \includegraphics[width=0.65\linewidth]{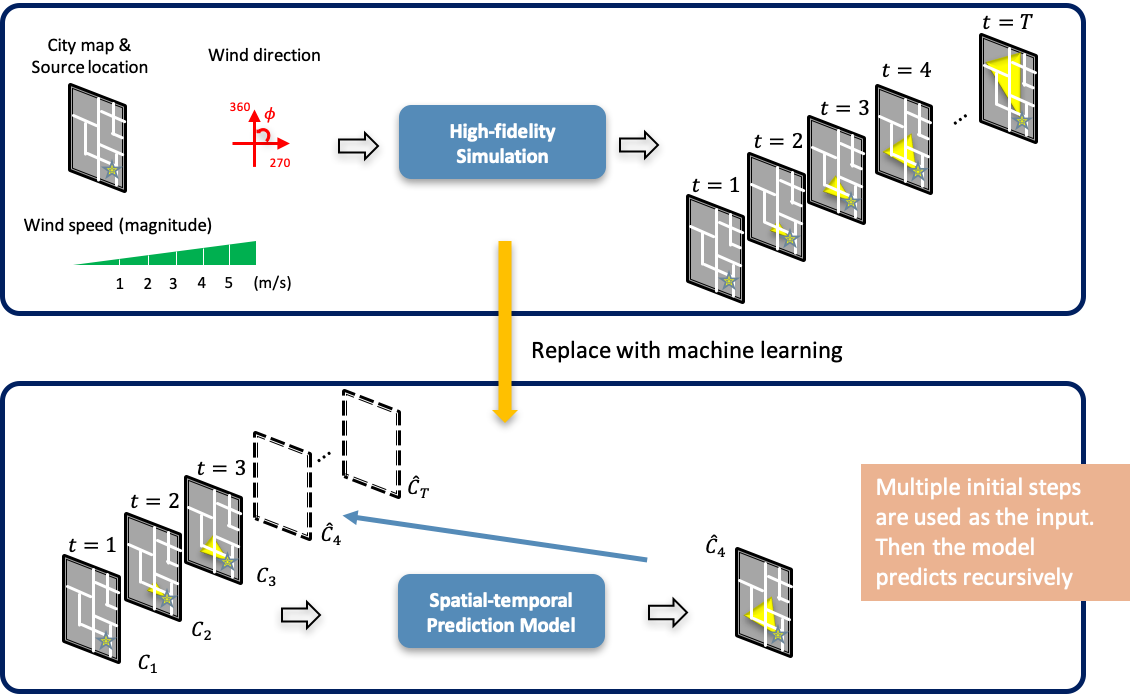}
    \caption{Schematic of the proposed workflow. The spatiotemporal ML model is trained and tested to replace part of the expensive high-fidelity physics-based model, allowing for real-time predictions of plume evolution from a few initial observations}
    \label{Fig:Overview}
\end{figure*}
Generating high spatial resolution predictions using high-fidelity CFD models can be infeasible for real-time predictions. Data-driven machine learning (ML) models can speed up this process while preserving the desired accuracy at a reduced cost \cite{atmos12080953, acp-5-1505-2005, acp-17-13521-2017}. Figure \ref{Fig:Overview} shows a schematic of the plume prediction workflow with and without the incorporation of ML models. The upper part of the figure shows the traditional workflow where initial conditions are used as inputs for a computationally expensive physics-based simulation to obtain high-fidelity spatiotemporal predictions. The lower part of the figure shows how the traditional method can be modified by adding a pre-trained ML model that can rapidly and accurately predict a comparable spatiotemporal prediction given a few initial observations of a previously unseen plume.
To make meaningful predictions on unseen data, ML models need to learn complex dependencies. For the considered plume dispersion problem, we identified four challenges: (i) real-time, physics-based simulation data is computationally expensive; (ii) plume evolution in urban areas can be affected by many factors; (iii) buildings can cause discontinuities in plumes that are difficult to predict; and (iv) ML models should be able to forecast plume evolution forward in time as far as possible with limited initial observations to anticipate future risks and consequences.

\begin{figure}[ht]
    \centering
    \includegraphics[width=0.5\linewidth]{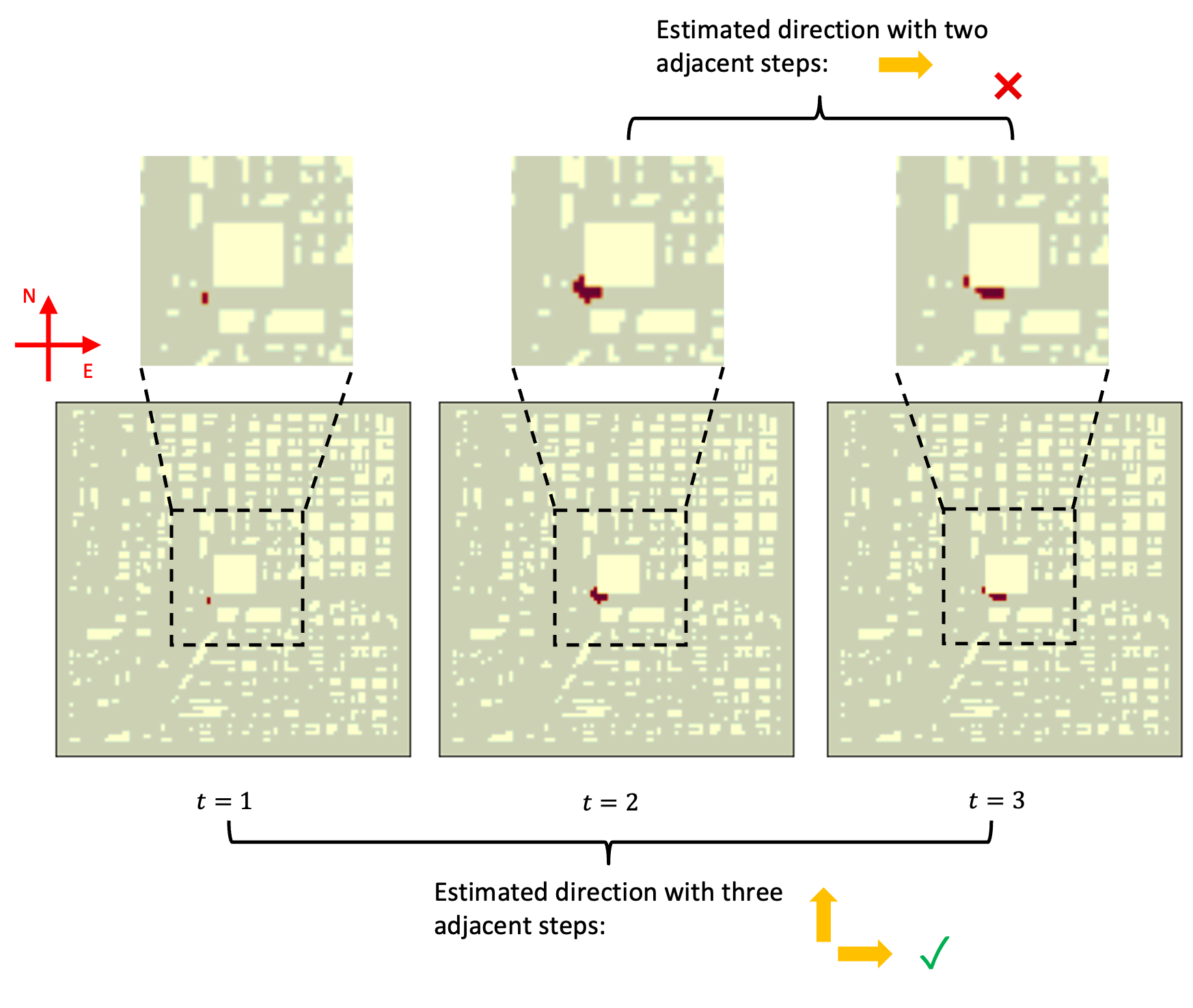}
    \caption{Example of the difficulties when estimating the dispersion process within two adjacent steps. The dispersion process happens simultaneously in multiple directions if spatial discontinuities (such as buildings) exist. Thus, considering enough early time steps is crucial when training the ML model}
    \label{Fig:Structure_motivation}
\end{figure}

To illustrate challenge (iii) above, Figure \ref{Fig:Structure_motivation} shows a spatial discontinuity that can occur when a plume encounters a building in a city. The figure displays a top-down view of the first three time steps of an atmospheric dispersion simulation (see Section \ref{sec:HFS} for simulation details). The plume is shown in red, and the yellow pixels represent buildings. If only influenced by wind, the plume should disperse toward the northeast. However, the blocking effect of the large central building forces the contaminant to disperse simultaneously along the east-west and north-south directions. When spatial discontinuities like this are present, explicitly considering only two previous time steps during ML training is insufficient. For example, in Figure \ref{Fig:Structure_motivation}, the progression from $t=2$ to $t=3$ suggests the dispersion is mainly in the east-west direction, which incorrectly leads the ML model to predict that the plume evolves only along this single direction. This issue motivates the inclusion of additional temporal observations in the training sequence (in this case, $t=1$, $t=2$, and $t=3$) when learning plume evolution behavior in urban areas.

To tackle these challenges, we propose a spatiotemporal deep-learning-based model to learn plume evolution in urban environments, which we will refer to as ST-GasNet for short. The contributions of ST-GasNet can be summarized into three aspects:

\begin{enumerate}
\item ST-GasNet can overcome spatial discontinuities, such as those caused by urban structures, by explicitly capturing both first-order and second-order information flows. This architecture is inspired by the presence of the second-order term in the advection-diffusion differential equation (as shown in Equation \ref{eq:AdvDiff}), which is widely adopted to model plume dispersion (as introduced in Section \ref{sec:HFS}).
\item ST-GasNet's ability to explicitly capture both first-order and second-order information flows enables it to more accurately model the complex dependencies between plume evolution and various environmental factors, such as wind direction, temperature, and topography, and to improve multi-step prediction even with limited data.
\item We propose a novel recurrent unit, ST-LSTM++, as the basic building module of ST-GasNet. It fuses relevant features from both first-order and second-order information flows and stores them in different memory states, which allows it to maintain a more comprehensive representation of the spatiotemporal patterns in the input data.
\end{enumerate}

The rest of this paper is organized as follows. Section \ref{sec:review} provides an overview of related research from the perspectives of plume dispersion prediction and general spatiotemporal prediction. In Section \ref{sec:preliminary}, we provide a detailed explanation of the simulation methods and ST-GasNet, including their key features and components. Section \ref{sec:method} presents the methodology used in ST-GasNet, including a clear and detailed explanation of the approach. In Section \ref{sec:results}, we present the results of the method and discuss their implications in detail. Section \ref{sec:conclusion} provides a summary of the key findings and implications of the study, as well as suggestions for future research or improvements to the method.

\section{Related Work} \label{sec:review}

This section reviews the related work from two perspectives. The first section reviews the data-driven ML models previously used to predict air quality and plume evolution, and the second section reviews general spatiotemporal prediction methods in related domains.

\subsection{Predicting Air Quality Using Data-driven Models}

Predicting urban air quality using data-driven methods is an active research area. Existing air quality prediction methods primarily focus on predicting the background levels of common pollutants in the atmosphere, such as $\text{NO}_{2}$ and $\text{PM}_{2.5}$. For instance, Zheng et al. proposed a semi-supervised learning approach that consisted of one spatial classifier and one temporal classifier to capture both spatial and temporal features in predicting $\text{PM}_{2.5}$ concentrations \cite{Zheng2013UAirWU}. Zheng et al. developed a data-driven method to predict the readings of air quality monitoring stations measuring six air pollutants \cite{10.1145/2783258.2788573}. Additionally, Qi et al. presented a unified framework called Deep Air Learning, which fused feature selection and semi-supervised learning to interpolate, predict, and analyze air quality data features\cite{8333777}. Lin et al. implemented a deep neural network (DNN)-based approach that fused environmental factors to predict air quality at individual monitoring stations \cite{10.1145/3274895.3274907}. Moreover, Wang proposed a deep learning model, HazeNet, to predict severe haze events in two megacities using time-sequential regional maps as inputs \cite{acp-21-13149-2021}. Li et al. developed an encoder-decoder full residual DNN to estimate spatiotemporal $\text{PM}{2.5}$ concentrations \cite{li2020encoder}. Gu et al. proposed a photograph-based monitoring model that uses a novel wide DNN to estimate real-time $\text{PM}_{2.5}$ concentrations \cite{gu2021pm}. Furthermore, Du et al. presented a hybrid deep learning framework that combined a Convolutional Neural Network (CNN) with a Bi-directional Long Short-term Memory (Bi-LSTM) network to predict multivariate time series data for air quality \cite{8907358}. Finally, Zhang et al. extended the prediction of air pollutants into a three-dimensional region using a multimodal fusion network \cite{9382281}.

Although closely related, the data-driven air quality literature cannot be applied directly to toxic plume evolution for several reasons. First,  background air pollutants commonly found in the atmosphere can be transported and mixed over long distances. The current methods are therefore primarily designed for predicting long-term air quality (over a period of days to months) rather than rapid dispersion from individual releases (over a period of minutes or hours). Second, there is more data available for training air quality machine learning models than toxic plume models. Lastly, air quality monitoring stations are often sparsely distributed across urban areas, which means that existing methods mainly focus on predicting future air quality at individual monitoring stations rather than predicting the full spatiotemporal pattern across the entire region.

\subsection{Spatiotemporal Prediction in Related Domains}
In our previous work, we demonstrated the effectiveness of deep convolutional autoencoder-based approaches driven by color images of spatial deposition in predicting new patterns for any initial source location and wind boundary condition \cite{Giselle2021,fernandez-godino_predicting_2023}. In this study, we expand upon our work to include spatiotemporal predictions. As depicted in Figure \ref{Fig:Overview}, our objective is to learn spatiotemporal dependencies from initial observations and recursively predict plume evolution. Spatiotemporal predictions have been extensively studied in related fields, such as predicting video frames \cite{8099713, wang2021predrnn}, traffic conditions \cite{10.5555/3298239.3298479, li2018diffusion}, and material degradation \cite{SCHWARZER2019322, wang2021}.

CNNs and their variants were initially proposed to learn spatiotemporal dependencies from historical data and to make predictions. \cite{DBLP:conf/iccv/TranBFTP15} showed how to use three-dimensional CNNs to learn spatiotemporal dependencies from videos for action recognition. \cite{Zhang2016DNNbasedPM} stacked observations of city traffic at different time steps into a tensor and designed a CNN-based model (DeepST) for learning spatiotemporal dependencies. The spatiotemporal Residual Network or ST-ResNet \cite{10.5555/3298239.3298479} improved the performance of DeepST by incorporating more information, such as weather, and extracting crucial features with a deeper network. In spatiotemporal predictions, CNNs are used as follows: (i) time series matrices are stored in a tensor, and then (ii) the CNN learns the spatial features at each time step and the temporal features among different time steps. However, due to model capacity limitations, CNN-based models have not performed well in tasks with complex spatiotemporal dependencies, such as video prediction.

The Long Short-Term Memory (LSTM) network and its variants were introduced to strengthen a model's ability to learn more complex temporal dependencies. A convolutional LSTM (ConvLSTM) network is a regular LSTM network where the gates are replaced with convolution operations \cite{NIPS2015_07563a3f}. This design allows the convolution operation to work as a spatial feature extractor at each time step and a temporal feature extractor that learns relationships from consecutive time series vectors. Some recent examples of environmental science applications of ConvLSTMs include the work of \cite{sinha2022week} on solar irradiance forecasting and the work of \cite{obakrim2023learning} on a two-stage model based on CNNs and LSTMs to learn the underlying spatiotemporal structure of the relationship between wind and ocean waves. The Predictive Recurrent Neural Network model (PredRNN) augments the ConvLSTM network with information flows along the zigzag route, which is a deep-in-time path of memory state transitions \cite{wang2021predrnn}. However, current ConvLSTM-type networks only explicitly consider dependencies between two adjacent time steps, which might be misled by spatial discontinuity plumes in a city, as discussed in Section \ref{sec:introduction}. This limitation is also validated by the experimental results of the baseline method shown in Section \ref{sec:results}.

\section{Preliminaries}
\label{sec:preliminary}
This section introduces two critical components of the proposed methodology ST-GasNet. Firstly, the high-fidelity physics-based simulation model, Aeolus~\cite{gowardhan2021large}, is utilized to generate the training data. Secondly, the state-of-the-art PredRNN \cite{wang2021predrnn} architecture, widely recognized for its superior spatiotemporal prediction performance, is employed in this study. Both Aeolus and PredRNN are integral to the development of ST-GasNet, which is discussed in detail in Section~\ref{sec:method}.

\subsection{Physical Model and High-Fidelity Simulation}
\label{sec:HFS}
\begin{figure}[ht]
    \includegraphics[width=0.5\linewidth]{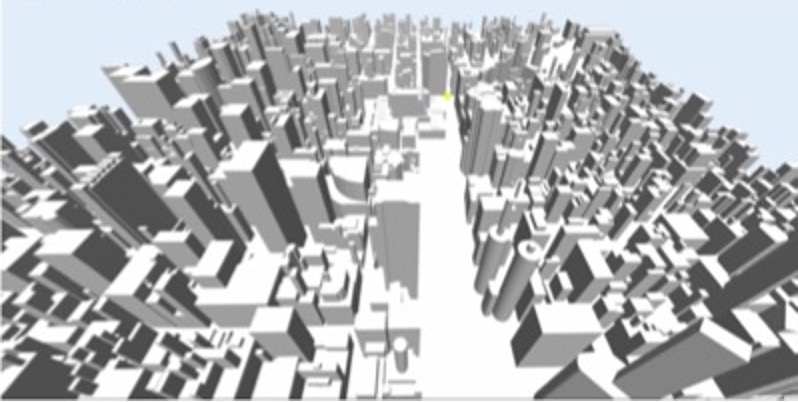}
    \centering 
    \caption{Aerial image of the Aeolus urban grid used for this problem}
    \label{Fig:Buildings}
\end{figure}

In this study, we used the three-dimensional computational fluid dynamics (CFD) code Aeolus to simulate toxic plumes. A three-dimensional Cartesian grid with coordinates $(x,y,z)$ is established in Aeolus (see schematic illustration in Figure~\ref{Fig:Buildings}). Here, $x$ and $y$ denote the east-west and north-south dimensions respectively, while $z$ represents the vertical dimension. The size of the urban domain is $2~\mathrm{km}$ in both the $x$- and $y$-directions and $0.4~\mathrm{km}$ in the $z$-direction. The grid resolution is set to $\Delta x = \Delta y = \Delta z = 4~\mathrm{m}$, resulting in a total of 500 cells in the east-west and north-south dimensions and 100 cells in the vertical dimension. The release of contaminants was simulated for 5 minutes from the center of the grid at ground level using one million Lagrangian particles. The simulations covered an hour of release with a time step of 0.5 seconds. To obtain a two-dimensional representation of the data for training an ML model, we considered only the concentration data at the first cell above the ground and converted the concentration values to binary values. A binary value of 1 was assigned to cells with contaminants and 0 to those without contaminants. For training the ML model, we used 50 time steps per simulation, with a time interval of 36 seconds between the steps. Therefore, a total of 30 minutes was modeled.

Aeolus is capable of predicting the transport and dispersion of materials in complex terrain and urban areas. For our simulations, we used Aeolus' Large Eddy Simulation (LES) option, which was validated in a previous study \cite{gowardhan2021large}. Aeolus predicts the dispersion of contaminants by solving the three-dimensional, incompressible, advection-diffusion equation with sources and sinks using a Lagrangian framework \cite{durbin1983nasa}. Further information can be found in the previous work \cite{gowardhan2021large}. The incompressible, three-dimensional Navier-Stokes equations are solved on a staggered mesh using a finite volume technique in Aeolus. The advective terms are discretized using a third-order Quadratic Upstream Interpolation for Convective Kinematics (QUICK) scheme \cite{leonard1979stable}, while the diffusive terms are discretized using a second-order central difference scheme. We used an accurate second-order Adams–Bashforth scheme for time integration.

\subsection{Predictive Recurrent Neural Network}

\begin{figure}[ht]
    \includegraphics[width=0.4\linewidth]{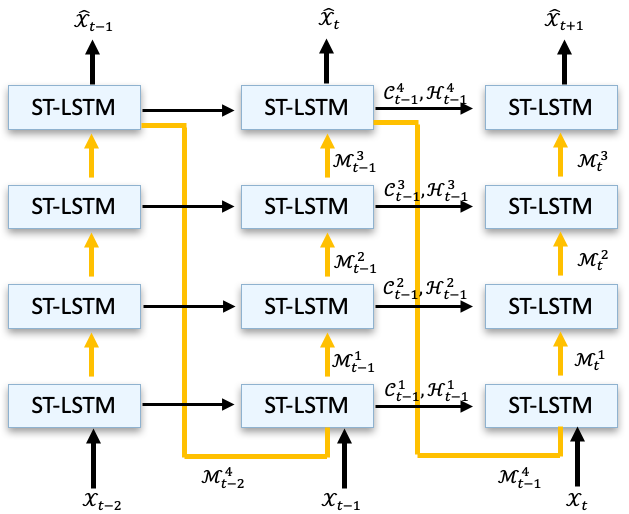}
    \centering 
    \caption{Structure of PredRNN. Arrowed lines in yellow and gray demonstrate information flow within PredRNN. Figure is adjusted from \cite{wang2021predrnn}}
    \label{Fig:PredRNN_Structure}
\end{figure}

The Predictive Recurrent Neural Network (PredRNN) model \cite{wang2021predrnn} is a state-of-the-art method for spatiotemporal and video prediction. The model's structure is illustrated in Figure \ref{Fig:PredRNN_Structure}. PredRNN comprises a four-layer model designed to capture the spatial and temporal dependencies of the input sequence simultaneously. Unlike pure temporal prediction models, such as the Long Short-Term Memory (LSTM) model \cite{HochSchm97}, which takes scalars or vectors as input, PredRNN takes a sequence of tensors as input. Compared with the ConvLSTM model \cite{NIPS2015_07563a3f}, PredRNN captures more complex spatiotemporal dependencies \cite{wang2021predrnn}. There are two major information flows in the model: the Horizon-flow, which propagates horizontally between two adjacent time steps (black arrows), and the Zigzag-flow, which propagates vertically (bottom to top) within each time step and then feeds into the bottom layer of the next time step (yellow lines).

The primary module of PredRNN is the spatiotemporal LSTM (ST-LSTM) \cite{wang2021predrnn}, which is shown in Figure \ref{Fig:PredRNN_Structure}. This module uses the following operations:
\begin{align}
g_{t} &= \text{tanh}(W_{xg} * \mathcal{X}_{t} + W_{hg} * \mathcal{H}_{t-1}^{l}),
\notag\\
i_{t} &= \sigma(W_{xi} * \mathcal{X}_{t} + W_{hi} * \mathcal{H}^{l}_{t-1}),
\notag\\
f_{t} &= \sigma(W_{xf} * \mathcal{X}_{t} + W_{hf} * \mathcal{H}^{l}_{t-1}),
\label{Eq:3-1.1}
\end{align}
their derivatives:
\begin{align}
g_{t}^{'} &= \text{tanh}(W_{xg}^{'} * \mathcal{X}_{t} + W_{mg} * \mathcal{M}_{t}^{l-1}),
\notag\\
i_{t}^{'} &= \sigma(W_{xi}^{'} * \mathcal{X}_{t} + W_{mi} * \mathcal{M}^{l-1}_{t}),
\notag\\
f_{t}^{'} &= \sigma(W_{xf}^{'} * \mathcal{X}_{t} + W_{mf} * \mathcal{M}^{l-1}_{t}),
\label{Eq:3-1.2}
\end{align}
and the memory cells and outputs:
\begin{align}
\mathcal{C}_{t}^{l} &= f_{t} \odot \mathcal{C}_{t-1}^{l} + i_{t} \odot g_{t},
\notag\\
\mathcal{M}_{t}^{l} &= f_{t}^{'} \odot \mathcal{M}_{t}^{l-1} + i_{t}^{'} \odot g_{t}^{'},
\notag\\
o_{t} &= \sigma(W_{xo} * \mathcal{X}_{t} + W_{ho} * \mathcal{H}_{t-1}^{l} 
\notag\\
&+ W_{co} * \mathcal{C}_{t}^{l} + W_{mo} * \mathcal{M}_{t}^{l}),
\notag\\
\mathcal{H}_{t}^{l} &= o_{t} \odot \text{tanh}(W_{1 \times 1} * [\mathcal{C}_{t}^{l}, \mathcal{M}_{t}^{l}]),
\label{Eq:3-1.3}
\end{align}
where $*$ denotes the convolution operation \cite{NIPS2012_c399862d}, $\odot$ denotes the Hadamard product \cite{0521386322}, $\sigma$ is the Sigmoid activation function \cite{10.1007/3-540-59497-3_175}, $\text{tanh}$ is the Hyperbolic Tangent activation function \cite{nwankpa2018activation}, $W_{\cdot \cdot}$ denotes the weights of convolutional kernels, $o_{t}$ is the output gate which fuses all the information contained in the current module, $\mathcal{X}_{t}$ is the input tensor at time step $t$, and $\mathcal{H}_{t}^{l}$ is the hidden state output. This module keeps the important information from the output gate and updates the features in the Horizon flow. Similar to the structure of the Convolutional LSTM model (ConvLSTM) \cite{NIPS2015_07563a3f}, the input gate $i_{t}$, forget gate $f_{t}$, and input-modulation gate $g_{t}$ together update the temporal memory $\mathcal{C}_{t}^{l}$. The spatiotemporal memory, $\mathcal{M}_{t}^{l}$, is newly introduced in PredRNN, which is updated by another set of gates $i_{t}^{'}, f_{t}^{'}, g_{t}^{'}$. 

It is worth noting that each ST-LSTM module has two memories, namely the temporal memory $\mathcal{C}_{t}^{l}$ and the spatiotemporal memory $\mathcal{M}_{t}^{l}$. The temporal memory is updated by fusing the features in the current observation $\mathcal{X}_{t}$ with the Horizon-flow $\mathcal{H}_{t-1}^{l}$ from the previous time step. The spatiotemporal memory is updated using the current observation $\mathcal{X}_{t}$ and the Zigzag-flow $\mathcal{M}_{t}^{l-1}$ from the previous module. The Zigzag-flow is then updated using the latest $\mathcal{M}_{t}^{l}$ from the current module and propagates to the following module. The Horizon-flow is updated by preserving relevant information in the module output based on the current temporal and spatiotemporal memories.

Compared with ConvLSTM, PredRNN adds a separate Zigzag-flow and a spatiotemporal memory, which provides two advantages for the model. First, it increases the model depth along the route of information flow, as depicted by the yellow arrows in Figure \ref{Fig:PredRNN_Structure}. Second, the extra spatiotemporal memory introduces additional parameters into each module, enabling it to capture more complex spatiotemporal dependencies.

However, there are still limitations in PredRNN's design. As shown in Equations \eqref{Eq:3-1.1}, \eqref{Eq:3-1.2} and \eqref{Eq:3-1.3}, when updating the information flows $\mathcal{H}_{t}^{l}$ and $\mathcal{M}_{t}^{l}$ at the current time step, only the information from the current and previous time steps is explicitly considered. Thus, when adapting PredRNN to predict plume evolution, the discontinuity in the spatial data may obscure the pattern of the evolution process and potentially mislead the model, as shown in Figure \ref{Fig:Structure_motivation}.

\section{Methodology} 
\label{sec:method}

To address the challenges in predicting plume evolution and bridge existing research gaps, we propose a novel spatiotemporal deep learning model called ST-GasNet. This model incorporates a specially designed "spatiotemporal LSTM ++" (ST-LSTM++) as its basic module. In this section, we present the two problem formulations considered in this work before introducing ST-GasNet in detail.

\subsection{Problem Formulation}

\begin{figure}[ht]
    \includegraphics[width=0.7\linewidth]{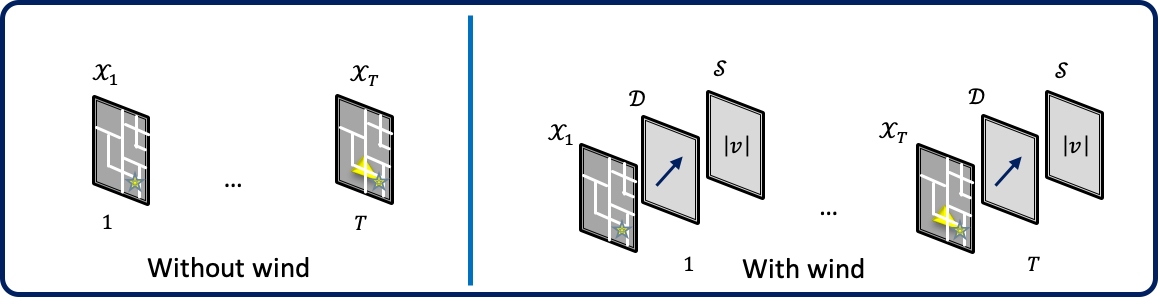}
    \centering 
    \caption{Two formulations of the problem. \textbf{Left}: Problem formulation without any wind information; \textbf{Right}: Problem formulation with wind information}
    \label{Fig:Formulations}
\end{figure}

Suppose we aim to predict the evolution of a plume; this problem has two possible formulations. One assumes that only historical observations of toxic plumes are available. The model needs to learn the evolution process from the historical data and then predict the future. The left side of Figure \ref{Fig:Formulations} shows the model input in this case, and the problem is formulated as:
\begin{align}
\max_{\theta} \sum_{\mathcal{T}}P_{\theta}(\mathcal{X}_{T+1:T+k} | \mathcal{X}_{1:T}),
\label{Eq:4-1}
\end{align}
where $P_{\theta}$ represents the probability of generating the true future observations; $\mathcal{T}$ is the set of training data containing all the pairs of $(\mathcal{X}_{1:T}, \mathcal{X}_{T+1:T+k})$; $\mathcal{X}_{1:T}$ represents a sequence of observations from time step $1$ to $T$, and $\mathcal{X}_{T+1:T+k}$ represents the following unobserved $k$ steps; $\mathcal{X}_{t}$ is a tensor with the shape of $N \times M \times 1$ representing the spatial distribution of toxic plumes over the monitored $N\times M$ grid space at time step $t$, which is referred to as one observation; and $\theta$ are the trainable parameters in ST-GasNet. The objective of this formulation is to find the values of parameters $\theta$ that maximize the probability of generating the true spatial distributions in the following $k$ steps given the observed data.

\label{sec:formulation}

An alternative formulation is to include environmental factors as part of the model input. In Section \ref{sec:HFS}, it was shown that the speed and direction of wind velocity are the most significant factors affecting plume evolution. To incorporate these factors into the model, we can modify the problem formulation as shown in the diagram on the right in Figure \ref{Fig:Formulations}. Specifically, we aim to maximize the probability of generating accurate future spatial distributions of toxic plumes given the current observations and environmental factors. The modified formulation is expressed as follows:
\begin{align}
\max_{\theta} \sum_{\mathcal{T}}P_{\theta}(\mathcal{X}_{T+1:T+k} | \mathcal{X}_{1:T}, \mathcal{D}, \mathcal{S}),
\label{Eq:4-2}
\end{align}
here, $\mathcal{D}$ represents the direction of wind over a monitored grid space, with each point represented by a unit vector over horizontal dimensions $x,y$. $\mathcal{S}$ represents the speed of wind. Both tensors have the shape of $N\times M$, where $N$ and $M$ represent the dimensions of the grid. By including the wind direction and wind speed, we explicitly include the driving factor that impacts the direction and velocity of the dispersion process. Therefore, the accuracy of predictions for future plume distributions can be improved.

It is important to acknowledge that although the direction and speed of the wind provide additional information to benefit the prediction, wind information may be partially available in the monitored region and obtainable only near monitoring stations. Thus, the more challenging problem formulated in Equation \eqref{Eq:4-1}, which takes into account these limitations, is closer to a real-world scenario.

\subsection{ST-GasNet}
\begin{figure*}[!ht]
    \includegraphics[width=0.8\linewidth]{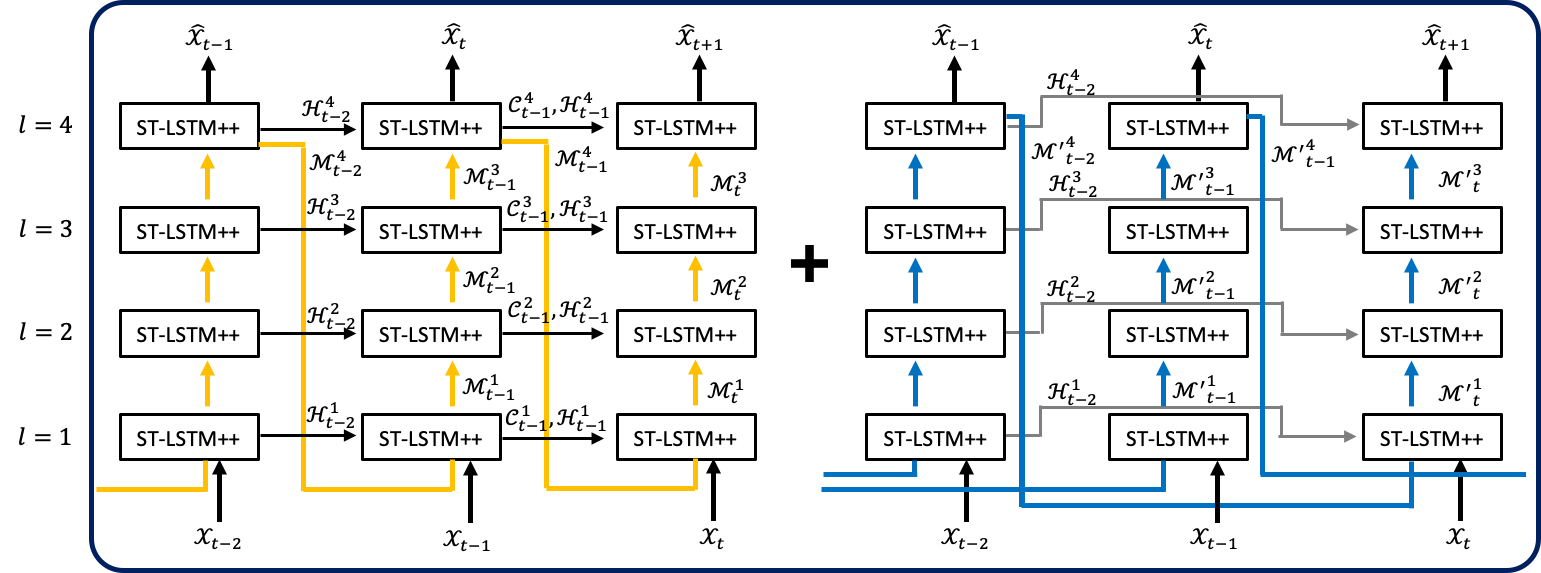}
    \centering 
    \caption{Structure of ST-GasNet. \textbf{Left}: First-order information flow (in yellow and black); \textbf{Right}: Second-order information flow (in blue and gray)}
    \label{Fig:GasNet_Structure}
\end{figure*}

As shown in Figure \ref{Fig:Structure_motivation}, capturing the trend of plume evolution can be difficult when spatial discontinuities are present and only two initial adjacent time steps are considered. To address this, we propose a model for learning more complex spatiotemporal dependencies. In the advection-diffusion Equation \eqref{eq:AdvDiff}, second-order spatial derivative terms drive the diffusion process. Thus, explicitly incorporating information from previous time steps introduces second-order information flow, which helps capture the trend of the dispersion process. We apply both spatial and temporal discretizations to our problem. By combining the second-order information flow with observations from the current time step, our proposed ST-GasNet model is able to capture more complex spatiotemporal dependencies contained in the second-order difference, a discrete analogy to the second-order derivative. The structure of our proposed model is shown in Figure \ref{Fig:GasNet_Structure}, where the first-order (left) and second-order (right) information flows are shown separately for easier visualization.

\begin{figure}[ht]
    \includegraphics[width=0.7\linewidth]{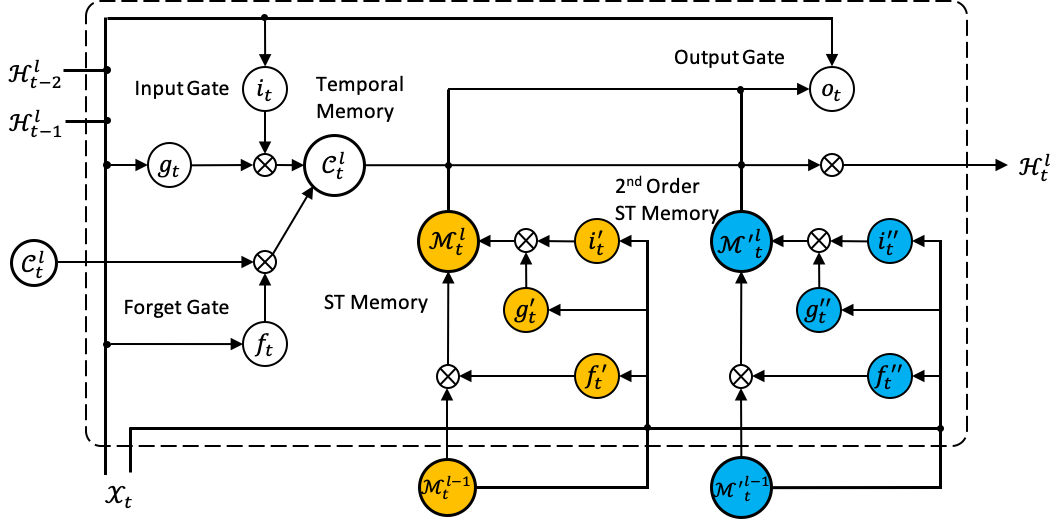}
    \centering 
    \caption{Structure of the Basic Module (ST-LSTM++) in ST-GasNet}
    \label{Fig:GasNet_Module}
\end{figure}

A novel building module, ST-LSTM++, has been specifically designed to exploit both first-order and second-order information flows simultaneously. The structure of ST-LSTM++ is illustrated in Figure \ref{Fig:GasNet_Module}, and the four routes of information flow are depicted in Figure \ref{Fig:GasNet_Structure} using arrowed lines of different colors (yellow, black, gray, and blue).

Let us consider the ST-LSTM++ module observation $\mathcal{X}{t}$ in the bottom layer $l=1$ at time step $t$. This observation contains the spatial features at the current time step and is inputted into the module. Additionally, we also include the first-order and second-order Horizon-flows, $\mathcal{H}_{t-1}^{1}$ and $\mathcal{H}_{t-2}^{1}$, respectively, as inputs. These contain spatiotemporal information from the two previous steps and are mainly used to update the temporal memory $\mathcal{C}_{t}^{l}$. The update operations for $\mathcal{C}_{t}^{l}$ can be expressed as
\begin{align}
g_{t} &= \text{tanh}(W_{xg} * \mathcal{X}_{t} + W_{hg} * \mathcal{H}_{t-1}^{l} + W_{hg}^{'} * \mathcal{H}_{t-2}^{l}),
\notag\\
i_{t} &= \sigma(W_{xi} * \mathcal{X}_{t} + W_{hi} * \mathcal{H}^{l}_{t-1}+ W_{hi}^{'} * \mathcal{H}_{t-2}^{l}),
\notag\\
f_{t} &= \sigma(W_{xf} * \mathcal{X}_{t} + W_{hf} * \mathcal{H}^{l}_{t-1}+ W_{hf}^{'} * \mathcal{H}_{t-2}^{l}),
\notag\\
\mathcal{C}_{t}^{l} &= f_{t} \odot \mathcal{C}_{t-1}^{l} + i_{t} \odot g_{t},
\label{Eq:4-3}
\end{align}
where the set of gates $g_{t}, i_{t}, f_{t}$ are specifically tailored to fuse the features from both $ \mathcal{H}_{t-1}^{l}$ and $\mathcal{H}_{t-2}^{l}$. This design allows the memory cell to store both first-order and second-order temporal dependencies in $\mathcal{C}_{t}^{l}$.

Besides the first- and second-order Zigzag flows, $\mathcal{M}_{t-1}^{L}$ and $\mathcal{M}_{t-2}^{'L}$ are also considered as inputs, where $L$ refers to the outputs from the top layer of the model. The model structure we used in this paper is shown in Figure \ref{Fig:GasNet_Structure}, in which $L = 4$. These inputs transport the high-level features from the top layer of the previous two time steps into the bottom layer of the current time step. This structure can both increase the depth of the model along with the propagation route and guide the model to store first-order and second-order features in spatiotemporal memories, which is expressed as:
\begin{align}
g_{t}^{'} &= \text{tanh}(W_{xg}^{'} * \mathcal{X}_{t} + W_{mg} * \mathcal{M}_{t}^{l-1}),
\notag\\
i_{t}^{'} &= \sigma(W_{xi}^{'} * \mathcal{X}_{t} + W_{mi} * \mathcal{M}^{l-1}_{t}),
\notag\\
f_{t}^{'} &= \sigma(W_{xf}^{'} * \mathcal{X}_{t} + W_{mf} * \mathcal{M}^{l-1}_{t}),
\notag\\
\mathcal{M}_{t}^{l} &= f_{t}^{'} \odot \mathcal{M}_{t}^{l-1} + i_{t}^{'} \odot g_{t}^{'},
\notag\\
g_{t}^{''} &= \text{tanh}(W_{xg}^{'} * \mathcal{X}_{t} + W_{mg}^{'} * \mathcal{M}_{t}^{'l-1}),
\notag\\
i_{t}^{''} &= \sigma(W_{xi}^{'} * \mathcal{X}_{t} + W_{mi}^{'} * \mathcal{M}^{'l-1}_{t}),
\notag\\
f_{t}^{''} &= \sigma(W_{xf}^{'} * \mathcal{X}_{t} + W_{mf}^{'} * \mathcal{M}^{'l-1}_{t}),
\notag\\
\mathcal{M}_{t}^{'l} &= f_{t}^{''} \odot \mathcal{M}_{t}^{'l-1} + i_{t}^{''} \odot g_{t}^{''}.
\label{Eq:4-4}
\end{align}
Here, $\mathcal{M}_{t}^{l}, \mathcal{M}_{t}^{'l}$ are the first- and second-order spatiotemporal memories, respectively. We set $\mathcal{M}_{t}^{0} = \mathcal{M}_{t-1}^{L}$ and $\mathcal{M}_{t}^{'0} = \mathcal{M}_{t-2}^{'L}$ as the input of the bottom layer $l=1$ at time step $t$. Note that we introduce a separate set of gates, $g_{t}^{''}, i_{t}^{''}, f_{t}^{''}$, to learn the second-order spatiotemporal memory separately. Unlike temporal memory, spatiotemporal memory has two properties: (i) the zigzag route of spatiotemporal memory enables it to capture more complex and high-level features, and (ii) the zigzag route is much longer than the horizon route, especially when the number of layers is increased, making it more challenging to preserve informative features throughout propagation. Introducing a separate set of gates for the second-order spatiotemporal memory will strengthen the model's ability to capture more complex features and keep the informative features through propagation, further improving the prediction.

Finally, the output of each ST-LSTM++ module and the update of the Horizon-flow are given as
\begin{align}
o_{t} &= \sigma(W_{xo} * \mathcal{X}_{t} + W_{ho} * \mathcal{H}_{t-1}^{l} + W_{ho}^{'} * \mathcal{H}_{t-2}^{l} 
\notag\\ 
&+ W_{co} * \mathcal{C}_{t}^{l} + W_{mo} * \mathcal{M}_{t}^{l} + W_{mo}^{'} * \mathcal{M}_{t}^{'l}),
\notag\\
\mathcal{H}_{t}^{l} &= o_{t} \odot \text{tanh}(W_{1 \times 1} * [\mathcal{C}_{t}^{l}, \mathcal{M}_{t}^{l}, \mathcal{M}_{t}^{'l}]).
\label{Eq:4-5}
\end{align}

ST-GasNet offers two major advantages. First, the model is guided to learn the first- and second-order spatiotemporal dependencies, which matches the general form of the advection-diffusion equation (as shown in Equation \eqref{eq:AdvDiff}). Second, the ST-LSTM++ is specifically designed to capture the first- and second-order information flows simultaneously. These advantages can be extended to other spatiotemporal prediction tasks as well.

\subsection{Loss Function}

The design of the loss function is specifically tailored for the proposed ST-GasNet, which is
\begin{align}
\mathcal{L}(\mathcal{X}_{1:T}, \mathcal{X}_{T+1:T+k}) &= \frac{1}{T+k-1}\sum_{t=2}^{T+k}||\widehat{\mathcal{X}}_{t} - \mathcal{X}_{t}||_{2}^{2}
\notag\\
& + \sum_{t=1}^{T+k}\sum_{l=1}^{L}\sum_{n} \frac{\langle \Delta \mathcal{C}_{t}^{l},  \Delta \mathcal{M}_{t}^{l}\rangle^{(n)}}{||\Delta \mathcal{C}_{t}^{l}||_{2}^{(n)} \cdot ||\Delta \mathcal{M}_{t}^{l}||_{2}^{(n)}}
\notag\\
& + \sum_{t=1}^{T+k}\sum_{l=1}^{L}\sum_{n} \frac{\langle \Delta \mathcal{C}_{t}^{l},  \Delta \mathcal{M}_{t}^{'l}\rangle^{(n)}}{||\Delta \mathcal{C}_{t}^{l}||_{2}^{(n)} \cdot ||\Delta \mathcal{M}_{t}^{'l}||_{2}^{(n)}},
 \label{Eq:4-6}
\end{align}
where $\widehat{\mathcal{X}}_{t}$ is the predicted spatial distribution of toxic plumes at time step $t$, $||.||_{2}^{2}$ denotes the square of the $l_{2}$ norm, $\langle.,.\rangle^{(n)}$ denotes the cosine similarity along the $n^{th}$ channel of two tensors, $||.||_{2}^{(n)}$ denotes the $l_{2}$ norm of a tensor along its $n^{th}$ channel, and $\Delta \mathcal{C}_{t}^{l}$, $\Delta \mathcal{M}_{t}^{l}$, and $\Delta \mathcal{M}_{t}^{'l}$ denote the updated temporal and spatiotemporal memories in the ST-LSTM++ module at time step $t$ and layer $l$, respectively. The expressions for $\Delta \mathcal{C}_{t}^{l}$, $\Delta \mathcal{M}_{t}^{l}$, and $\Delta \mathcal{M}_{t}^{'l}$ are:
\begin{align}
\Delta \mathcal{C}_{t}^{l} &= i_{t} \odot g_{t},
\notag\\
\Delta \mathcal{M}_{t}^{l} &= i_{t}^{'} \odot g_{t}^{'},
\notag\\
\Delta \mathcal{M}_{t}^{'l} &= i_{t}^{''} \odot g_{t}^{''},
\label{Eq:4-7}
\end{align}
where $i_{t}, g_{t}$, $i_{t}^{'}, g_{t}^{'}$, and $i_{t}^{''}, g_{t}^{''}$ are three sets of gates controlling the memories updates in each ST-LSTM++ module and defined in Equations \eqref{Eq:4-3} and \eqref{Eq:4-4}. 

The proposed loss function has a two-fold intuition. The first term aims to minimize the difference between predicted and true observations. This approach ensures that the trained model generates the most likely future distributions of toxic plumes given the current observations. The second and third terms guide the model to store different features in the temporal memory and the first-order and second-order spatiotemporal memories, respectively. This approach ensures that the corresponding Horizon and Zigzag flows cover complementary informative features to describe the dispersion process of toxic plumes, thereby improving prediction accuracy. Note that we allow the first-order and second-order Zigzag flows to share some similarities by not forcing the first- and second-order spatiotemporal memories to be different. In summary, this loss function aims to guide ST-GasNet to capture as many informative spatiotemporal dependencies as possible to generate the most accurate predictions.

\section{Experiments and Results} 
\label{sec:results}
This section discusses further experiment details and compares the performances between ST-GasNet and the baseline method.

\subsection{Dataset Introduction}
\label{sec:data}
The ST-GasNet architecture is trained and tested on simulated toxic plumes influenced by different wind velocities in an urban domain. Each simulation uses a different inflow condition that supplies the wind speed and direction at the lateral boundary. For a given inflow condition, the CFD model calculates the winds internally in the domain. Inflow wind direction samples are available every 10 degrees in the interval [180$^{\circ}$, 270$^{\circ}$] (inflow from directions between south and west), and inflow wind speeds every 1~$m/s$ in the interval [1, 5]~$m/s$. Dispersion in this domain occurs quickly for these wind speeds. It takes no longer than about 30 minutes for plumes to traverse the grid.

% \begin{figure}[ht]
%     \includegraphics[width=0.5\linewidth]{images/wind_direction_magnitude.png}
%     \centering 
%     \caption{Range of wind possible velocity values (direction and magnitude) in simulations}
%     \label{Fig:wind_info}
% \end{figure}

Our goal is to predict the evolution of plumes immediately after their initial release. To achieve this, the simulations were output in 36-second intervals over a 30-minute span, resulting in 50 time steps per simulation. In addition, we limit our focus to two-dimensional evolution by integrating the plumes over the z direction. The urban region of interest covers a $2 km \times 2 km$ area, which is discretized into $100 \times 100$ cells, with each cell having 20-meter resolution in the x and y directions. To further simplify the dataset, plume concentrations are replaced with a binary value in each cell. A value of 1 indicates the presence of the contaminant, while 0 indicates its absence. It is important to note that the toxic plumes are discontinuously distributed in the region due to the presence of buildings.

The dataset contains $44$ sequences of observations ${\mathcal{X}^{i}, i = 1, ..., 44}$, each representing the evolution of a specific plume and stored in a tensor with a shape of $100 \times 100 \times 50$. We formulate each sequence into pairs of inputs and outputs to train and test the models, which we refer to as clips. Suppose each input consists of $T$ time steps and each output consists of $k$ time steps. A clip from sequence $i$ can be denoted as $(\mathcal{X}^{i}_{t_{0}:t_{0}+T}, \mathcal{X}^{i}_{t_{0}+T:t_{0}+T+k})$, where $t_{0}$ is the starting point. During training, both the input and output are available for the model, while during testing, only the input is available, and the model generates all $k$ steps of output. Multiple clips can be generated from each sequence of observations, determined by the stride value $s$, which indicates the interval between the starting point of two adjacent inputs. For example, if a clip is $(\mathcal{X}^{i}_{t_{0}:t_{0}+T}, \mathcal{X}^{i}_{t_{0}+T:t_{0}+T+k})$, the next clip is $(\mathcal{X}^{i}_{t_{0}+s:t_{0}+s+T}, \mathcal{X}^{i}_{t_{0}+s+T:t_{0}+s+T+k})$. Note that this architecture allows for overlap among different clips. The number of clips generated from each sequence during the training phase can be calculated as $\frac{50-(T+k)}{s} + 1$.

\subsection{Experiment Setup}
In our experiment, we set $T=5$, $k=15$, and $s=2$ during the training phase. We trained the model to predict plume evolution during the next $9$ minutes when given $3$ minutes of observations. During the testing phase, we focused on the model's prediction ability at an early stage because it is more challenging to capture the evolution trends (as shown in Figure \ref{Fig:Structure_motivation}), but it has the highest impact when serving as an early-warning system in practice. That is, if we can accurately predict the future pattern earlier, the decision-maker will have more time to react. Thus, we evaluated the performance of our model by examining the prediction results of the first clip of each sequence (predicting the next $9$ minutes with the initial $3$ minutes as the input).

We randomly selected 36 sequences as the training data and used the remaining eight simulations as the testing data. The direction of the wind velocity for sequence $i$ is denoted as $\mathcal{D}^{i}$, which is a tensor with the shape of $100 \times 100 \times 2$, and the speed of the wind velocity is denoted as $\mathcal{S}^{i}$, which is a tensor with the shape of $100 \times 100 \times 1$. The direction and speed represent the large-scale inflow, and each entry in $\mathcal{D}^{i}$ and $\mathcal{S}^{i}$ are expressed as $\left(\cos(2\pi - \phi^{i}), \sin(2\pi - \phi^{i})\right)$ and $|v^{i}|$, respectively. The combinations of angle $\phi^{i}$ and magnitude $|v^{i}|$ differ between simulations. As introduced in Section \ref{sec:formulation}, the wind information can be included as part of the model input. If the wind information is included, the observation from sequence $i$ at time step $t$ is augmented as $[\mathcal{X}_{t}^{i}, \mathcal{D}^{i}, \mathcal{S}^{i}]$, which is a tensor with the shape of $100 \times 100 \times 4$.

The performance of the proposed ST-GasNet model is assessed against PredRNN, which serves as the benchmark method. To evaluate the predicted observation against the real physics model observation at each time step, the precision and modified accuracy are selected as evaluation metrics. The precision and accuracy are defined as follows:
\begin{align}
\text{Precision}(\mathcal{X}_{t}^{i}, \widehat{\mathcal{X}}_{t}^{i}) &= \frac{\text{TP}}{\text{TP} + \text{FP}},
\notag\\
\text{Accuracy}(\mathcal{X}_{t}^{i}, \widehat{\mathcal{X}}_{t}^{i}) &= \frac{\text{TP} + \frac{\text{TN}}{4}}{\text{TP} + \text{FP} + \text{FN} + \frac{\text{TN}}{4}},
\label{Eq:5-1}
\end{align}
where $\text{TP}$ and $\text{FP}$ denote the overall number of grid cell entries in which the model correctly and incorrectly predicts the existence of toxic plumes, respectively, while $\text{TN}$ and $\text{FN}$ denote the overall number of grid cell entries in which the model correctly and incorrectly predicts the non-existence of toxic plumes, respectively.

Because of class imbalances, we modified the expression of accuracy to assign a lower weight for predicting the absence of toxic plumes correctly. As shown in Figure \ref{Fig:Structure_motivation}, toxic plumes occupy only a small region in the early stages. The majority of readings will be counted as $\text{TN}$, causing the maximum value of $\text{TP}$ to be significantly smaller than the maximum value of $\text{TN}$. In the scenario where the model predicts no contamination in any part of the monitored area, the accuracy will still appear high in the early stages because it is dominated by the value of $\text{TN}$. To address this, we divided the value of $\text{TN}$ uniformly by four to ensure that the maximum value of $\text{TP}$ was of a similar magnitude to the maximum value of the adjusted $\text{TN}$. We chose the value of four because we observed that toxic plumes would occupy, at most, a quarter of the monitored area (see Figure \ref{Fig:results_viz_7}).

\subsection{Results}

\begin{figure*}[ht]
    \includegraphics[width=\linewidth]{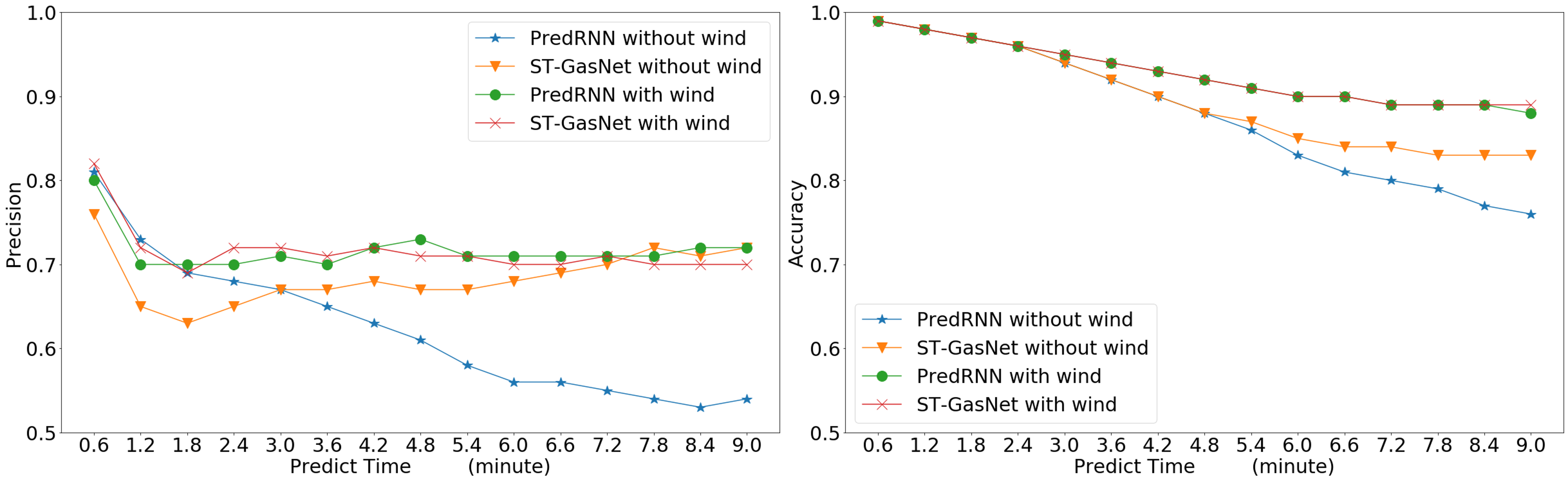}
    \centering 
    \caption{PredRNN and ST-GasNet performance metrics with or without wind information. \textbf{Left}: PredRNN and ST-GasNet precision with or without wind information (range = $[0,1]$); \textbf{Right}: PredRNN and ST-GasNet modified accuracy with or without wind information (range = $[0,1]$)}
    \label{Fig:eval_metric}
\end{figure*}

During the testing phase, we compared ST-GasNet and PredRNN to eight reserved sequences. Figure \ref{Fig:eval_metric} visualizes the average precision and accuracy over these sequences at each time step, while Tables \ref{table:5-1} and \ref{table:5-2} provide further details. Our numerical results reveal that: (i) ST-GasNet (orange triangle line) outperforms PredRNN (blue star line) in both precision and accuracy when wind information is not available for prediction, (ii) ST-GasNet (red cross line) and PredRNN (green dot line) have comparable performances when wind information is available, and (iii) ST-GasNet without wind information performs similarly to PredRNN with wind information. These findings demonstrate that the novel ST-GasNet structure successfully learns the direction and speed of dispersion processes from limited observations and maintains this information for multi-step prediction. Furthermore, our results suggest that incorporating wind information only marginally improves ST-GasNet's performance. This is beneficial for practical applications where accurate wind information may not be available for the entire monitored region. Overall, ST-GasNet appears to be a promising approach for gas dispersion prediction.

\begin{table*}[!ht]
    \centering 
    \caption{Performances comparison among different models using precision}
    \begin{tabular}{ccccccccccccccccc}
    \hline
    Formulation & Model & $t = 6$ & 7 & 8 & 9 & 10 & 11 & 12 & 13 & 14 & 15 & 16 & 17 & 18 & 19 & 20 \\
    \hline
    \multirow{2}{*}{Without Wind}& PredRNN & 0.81 & 0.73 & 0.69 & 0.68 & 0.67 & 0.65 & 0.63 & 0.61 & 0.58 & 0.56 & 0.56 & 0.55 & 0.54 & 0.53 & 0.54 \\
    & ST-GasNet & 0.76 & 0.65 & 0.63 & 0.65 & 0.67 & 0.67 & 0.68 & 0.67 & 0.67 & 0.68 & 0.69 & 0.70 & 0.72 & 0.71 & 0.72 \\
     \hline
    \multirow{2}{*}{With Wind}& PredRNN & 0.80 & 0.70 &  0.70 & 0.70 & 0.71 & 0.70 & 0.72 & 0.73 & 0.71 & 0.71 &  0.71 & 0.71 & 0.71 &  0.72 & 0.72 \\
    & ST-GasNet & 0.82 & 0.72 & 0.69 & 0.72 & 0.72 & 0.71 &  0.72 & 0.71 & 0.71 & 0.70 & 0.70 & 0.71 & 0.70 & 0.70 &  0.70 \\
    \hline
    \end{tabular}
    \label{table:5-1}
\end{table*}

\begin{table*}[!ht]
    \centering
    \caption{Performances comparison among different models using modified accuracy}
    \begin{tabular}{ccccccccccccccccc}
    \hline
    Formulation & Model & $t = 6$ & 7 & 8 & 9 & 10 & 11 & 12 & 13 & 14 & 15 & 16 & 17 & 18 & 19 & 20 \\
    \hline
    \multirow{2}{*}{Without Wind}& PredRNN & 0.99 & 0.98 & 0.97 & 0.96 & 0.94 & 0.92 & 0.90 & 0.88 & 0.86 & 0.83 & 0.81 & 0.80 & 0.79 & 0.77 & 0.76 \\
    & ST-GasNet & 0.99 & 0.98 & 0.97 & 0.96 & 0.94 & 0.92 & 0.90 & 0.88 & 0.87 & 0.85 & 0.84 & 0.84 & 0.83 & 0.83 & 0.83 \\
    \hline
    \multirow{2}{*}{With Wind}& PredRNN & 0.99 & 0.98 & 0.97 & 0.96 & 0.95 & 0.94 & 0.93 & 0.92 & 0.91 & 0.90 & 0.90 & 0.89 & 0.89 & 0.89 & 0.88 \\
    & ST-GasNet & 0.99 & 0.98 & 0.97 & 0.96 & 0.95 & 0.94 & 0.93 & 0.92 & 0.91 & 0.90& 0.90 & 0.89 & 0.89 & 0.89 & 0.89 \\
    \hline
    \end{tabular}
    \label{table:5-2}
\end{table*}

\begin{figure*}[ht]
    \includegraphics[width=0.85\linewidth]{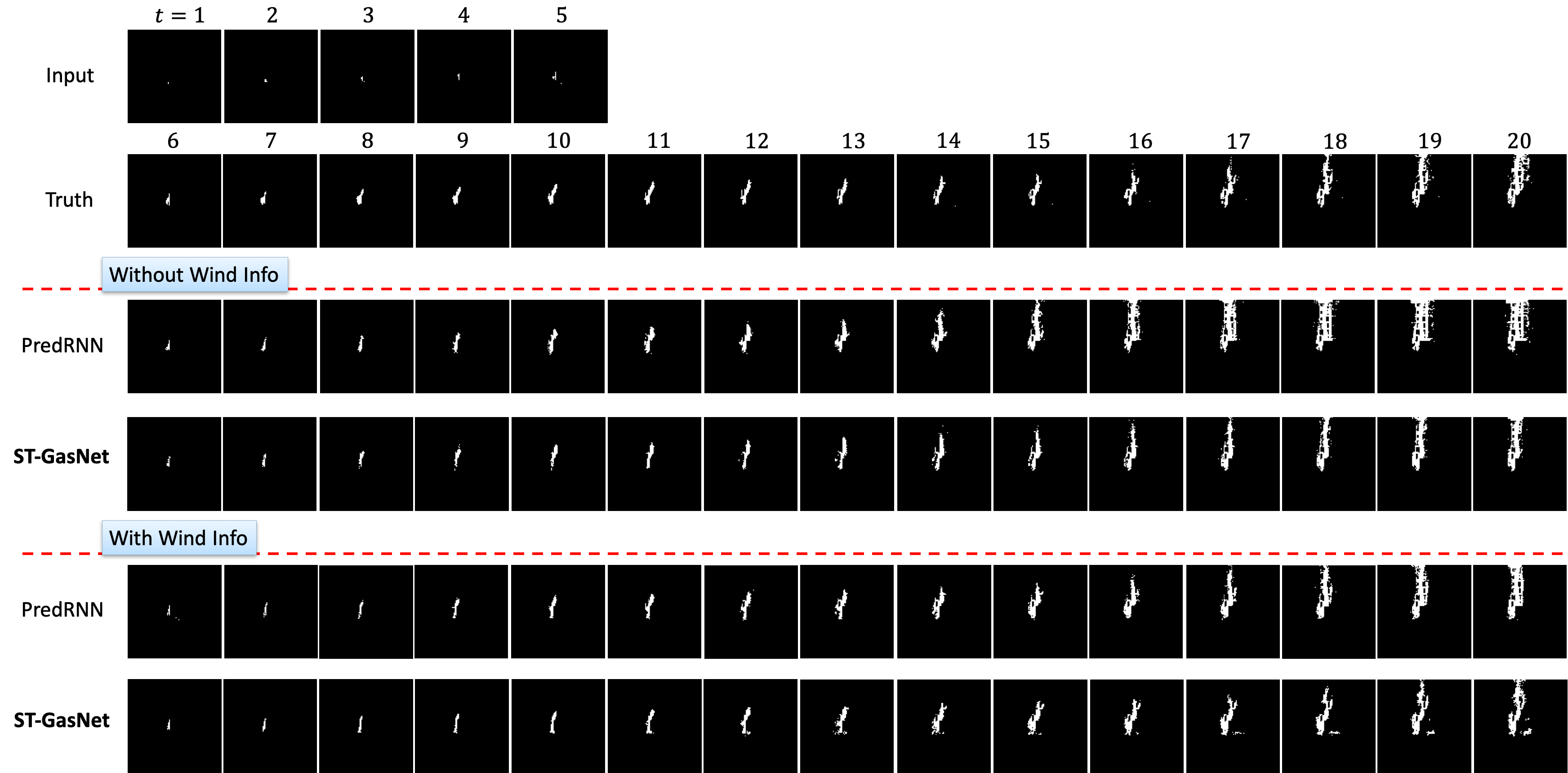}
    \centering 
    \caption{Test result 1 (randomly selected from the set of eight) on predicting plume evolution influenced by wind velocity with a direction of 200 degrees and a speed of 3 m/s. The results include the different formulations (with/without wind information) for the proposed ST-GasNet and the baseline method PredRNN}
    \label{Fig:results_viz_1}
\end{figure*}

\begin{figure*}[ht]
    \includegraphics[width=0.85\linewidth]{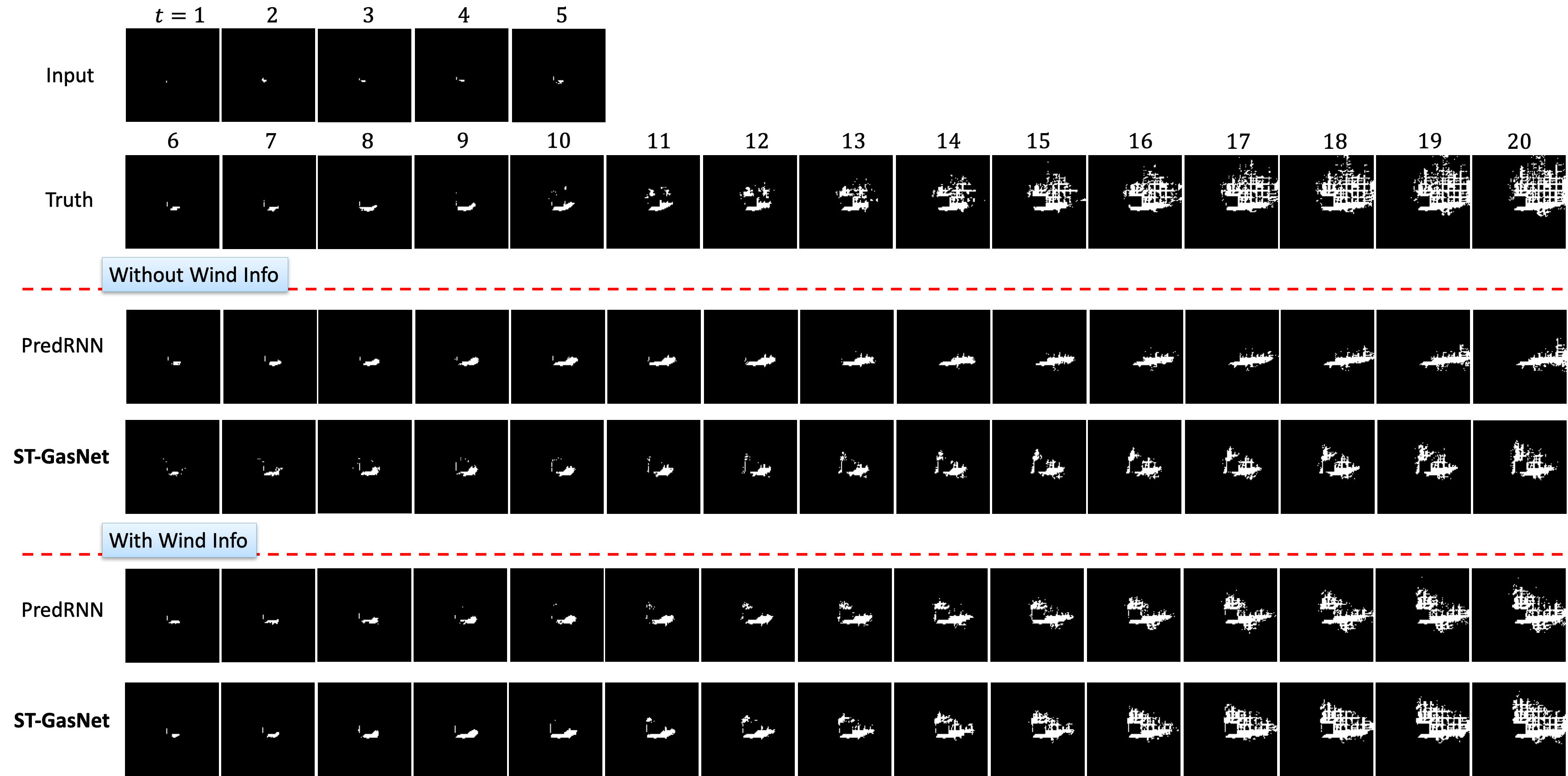}
    \centering 
    \caption{Test result 2 (randomly selected from the set of eight) on predicting plume evolution influenced by wind velocity with a direction of 240 degrees and a speed of 3 m/s. The results include the different formulations (with/without wind information) for the proposed ST-GasNet and the baseline method PredRNN}
    \label{Fig:results_viz_7}
\end{figure*}

During evaluation, PredRNN successfully predicted the outcome of five out of eight testing sequences when wind information was not provided, while ST-GasNet accurately predicted all eight sequences. Two testing sequences were randomly selected for visualization, as shown in Figure \ref{Fig:results_viz_1} and Figure \ref{Fig:results_viz_7}. In Figure \ref{Fig:results_viz_1}, both models predicted the future pattern of toxic plumes when wind information was not available, but PredRNN's predicted dispersion was faster than the ground truth. Specifically, the toxic plume reached the boundary of the monitored region at $t=15$ according to PredRNN, while the ground truth indicated it reached the boundary at $t=18$. ST-GasNet's prediction was more accurate, with the toxic plume reaching the boundary at $t=17$. When wind information was available ($340^{\circ}, 3 m/s$), as shown in the last row of Figure \ref{Fig:results_viz_1}, ST-GasNet predicted the direction of dispersion more accurately than PredRNN, with a slight inclination to the east instead of a strictly spreading towards the north.

Figure \ref{Fig:results_viz_7} displays the performance of the model on a separate testing sequence, which serves as the motivating example (shown in Figure \ref{Fig:Structure_motivation}). The results obtained without including wind information confirm the efficacy of ST-GasNet. The innovative structure of ST-GasNet enhances its ability to predict the future pattern of toxic plumes by explicitly incorporating second-order information flows. These findings are significant because wind information may not be fully accessible in practical scenarios, indicating that ST-GasNet can accurately capture the propagation process's direction and speed purely from limited historical observations and further exploit such information for improved predictions.

\section{Conclusion} \label{sec:conclusion} 

Predicting plume behavior in urban regions is a highly complex task due to the presence of obstacles, such as buildings, that impose discontinuities, and the need for accurate predictions often results in computationally prohibitive discretization. To address these challenges, we propose ST-GasNet, a novel spatiotemporal prediction model that explicitly learns the first- and second-order spatiotemporal dependencies from a limited set of observations to improve predictions. In this work, we evaluated the performance of ST-GasNet by predicting the dispersion patterns of 15 observations in the future, given the initial 5 time steps as input. Our results demonstrate that ST-GasNet successfully predicted the dispersion patterns of all the testing cases and outperforms the benchmark method (PredRNN) when comparing precision and modified accuracy. When evaluated on never-seen test cases, ST-GasNet achieved an accuracy of 90\% in the 9-minute window considered. The advantages of ST-GasNet can be summarized as follows: (i) it can successfully predict dispersion trends even without wind information, which is not always available everywhere in the monitored region, (ii) its design is inspired by the advection-diffusion equation to explicitly capture the first- and second-order information flows, allowing the model to learn complex spatiotemporal dependencies from both first- and second-order derivatives, and (iii) it is not only designed for predicting plume evolution but can also be applied to general spatiotemporal prediction tasks. 

The results from our experiments also demonstrate the practical impact of ST-GasNet in predicting plume evolution in urban regions. Given that wind information is usually unavailable in urban space, it is important to have efficient and accurate predictions of plume evolution only based on a few initial observations. Compared with the intensive computing cost of existing CFD-based simulation (several hours), the proposed ST-GasNet provides almost instant estimations on unseen scenarios (less than 1 minute of inference time), which makes it a valuable addition to emergency response planning and mitigation efforts.

\section{Data Availability}
The data used in this work is open-sourced and available upon request.
% % use section* for acknowledgment
% \ifCLASSOPTIONcompsoc
%   % The Computer Society usually uses the plural form
%   \section*{Acknowledgments}
    
% \else
%   % regular IEEE prefers the singular form
% \fi


\bibliographystyle{IEEEtran}
\bibliography{IEEEabrv,references}
\end{document}